\documentclass[11pt,a4paper]{article}
\usepackage[hyperref]{acl2021}
\usepackage{times}
\usepackage{latexsym}

\usepackage{microtype}

\usepackage{amsmath,amssymb}
\usepackage{mathtools}
\usepackage{multirow}
\usepackage[inline]{enumitem}
\usepackage[noend,noline,boxruled,commentsnumbered,linesnumbered,titlenumbered]{algorithm2e}

\SetCommentSty{mycommfont}

\usepackage{tikz}
\usepackage{tikz-qtree}
\usepackage{booktabs}
\usepackage[T1]{fontenc}

\aclfinalcopy % Uncomment this line for the final submission
\newcommand\scan{\textsc{SCAN}\xspace}
\newcommand\scansp{\textsc{SCAN-SP}\xspace}
\newcommand\clevr{\textsc{CLEVR}\xspace}
\newcommand\closure{\textsc{CLOSURE}\xspace}
\newcommand\geo{\textsc{GeoQuery}\xspace}
\newcommand\bb{\texttt{BERT-base}\xspace}
\newcommand\sbm{\textsc{SpanBasedSP}\xspace}
\newcommand\stos{\textsc{Seq2Seq}\xspace}
\newcommand\grammar{\textsc{grammar}\xspace}
\newcommand\etoe{\textsc{End2End}\xspace}
\newcommand\btos{\textsc{BERT2Seq}\xspace}
\newcommand\bart{\textsc{BART}\xspace}
\newcommand\bartb{\texttt{BART-base}\xspace}

\newcommand\iid{\textsc{iid}\xspace}

\newcommand\kbc{\ensuremath{\Sigma}\xspace}
\newcommand\s{\texttt{join}\xspace}
\newcommand\ns{\texttt{\ensuremath{\phi}}\xspace}
\newcommand\nsb{\texttt{$\boldsymbol\phi$}\xspace}
\newcommand\sent{\texttt{S}\xspace}

\definecolor{langcol}{HTML}{0D7315}
\definecolor{nodecol}{HTML}{8B0A35}
\definecolor{progcol}{HTML}{0151BD}
\newcommand\ttok[2]{{\textbf{\color{langcol}\emph{#1}}}\textsubscript{#2}}
\newcommand\tnode[1]{\textbf{\texttt{\color{nodecol}{#1}}}}
\newcommand\tprog[1]{\texttt{\color{progcol}{#1}}}
\newcommand\utt[1]{\emph{``#1''}}
\newcommand\comment[1]{}
\DeclareMathOperator*{\argmax}{argmax}

\title{Span-based Semantic Parsing for Compositional Generalization}

\author{Jonathan Herzig$^{1}$  ~~~~~~~~~~~~~~~~~~~~Jonathan Berant$^{1,2}$ \\
\mbox{}\\
$^1$Blavatnik School of Computer Science, Tel-Aviv University \\
$^2$Allen Institute for Artificial Intelligence \\
\small{\texttt{\{jonathan.herzig,joberant\}@cs.tau.ac.il}}
}

\date{}

\begin{document}
\maketitle
\begin{abstract}
  Despite the success of sequence-to-sequence (seq2seq) models in semantic parsing, recent work has shown that they fail in \emph{compositional generalization}, i.e., the ability to generalize to new structures built of components observed during training. In this work, we posit that a \emph{span-based} parser should lead to better compositional generalization. we propose \sbm, a parser that predicts a span tree over an input utterance, explicitly encoding how partial programs compose over spans in the input. \sbm extends \newcite{pasupat-etal-2019-span} to be comparable to seq2seq models by (i) training from programs, without access to gold trees, treating trees as latent variables, (ii) parsing a class of non-projective trees through an extension to standard CKY. On \geo, \scan and \closure datasets, \sbm performs similarly to strong seq2seq baselines on random splits, but dramatically improves performance compared to baselines on splits that require compositional generalization: from $61.0 \rightarrow 88.9$ average accuracy.
\end{abstract}

\section{Introduction}

The most dominant approach in recent years for semantic parsing, the task of mapping a natural language utterance to an executable program, has been based on sequence-to-sequence (seq2seq) models \cite[\emph{inter alia}]{jia2016recombination,dong-lapata-2016-language,wang-etal-2020-rat}. In these models, the output program is decoded step-by-step (autoregressively), using an attention mechanism that softly ties output tokens to the utterance.

\begin{figure*}[t]
\centering
% \footnotesize
\footnotesize
\begin{tikzpicture}
\tikzset{level distance=20pt}
\tikzset{frontier/.style={distance from root=140pt}}
\tikzset{every tree node/.style={align=center,anchor=north}}
\Tree [.\tnode{\s{}}:\;\tprog{capital(loc\_2(state(next\_to\_1(NY)))}
[.\tnode{\nsb{}} \edge[roof]; {\ttok{What}{1}\;\; \ttok{is}{2}\;\; \ttok{the}{3}} ] [.\tnode{\s{}}:\;\tprog{capital(loc\_2(state(next\_to\_1(NY)))} [.\tnode{capital} \ttok{capital}{4} ] [.\tnode{\s{}}:\;\tprog{loc\_2(state(next\_to\_1(NY)))} [.\tnode{loc\_2} \ttok{of}{5} ] [.\tnode{\s{}}:\;\tprog{state(next\_to\_1(NY))} [.\tnode{\s{}}:\;\tprog{state} [.\tnode{state} \ttok{states}{6} ] [.\tnode{\nsb{}} \ttok{that}{7} ] ] [.\tnode{\s{}}:\;\tprog{next\_to\_1(NY)} [.\tnode{stateid('new york')} \edge[roof]; {\ttok{New}{8}\;\; \ttok{York}{9}} ] [.\tnode{\s{}}:\;\tprog{next\_to\_1}  [.\tnode{next\_to\_1} \ttok{borders}{10} ] [.\tnode{\nsb{}} \ttok{?}{11} ] ] ] ] ] ] ]
\end{tikzpicture}
\caption{An example span tree. Nodes are annotated with categories (in bold). A node with a category \s over the span $(i,j)$, is annotated with its sub-program $z_{i:j}$. We abbreviate \texttt{stateid('new york')} to \texttt{NY}.}
\label{fig:tree_example}
\end{figure*}
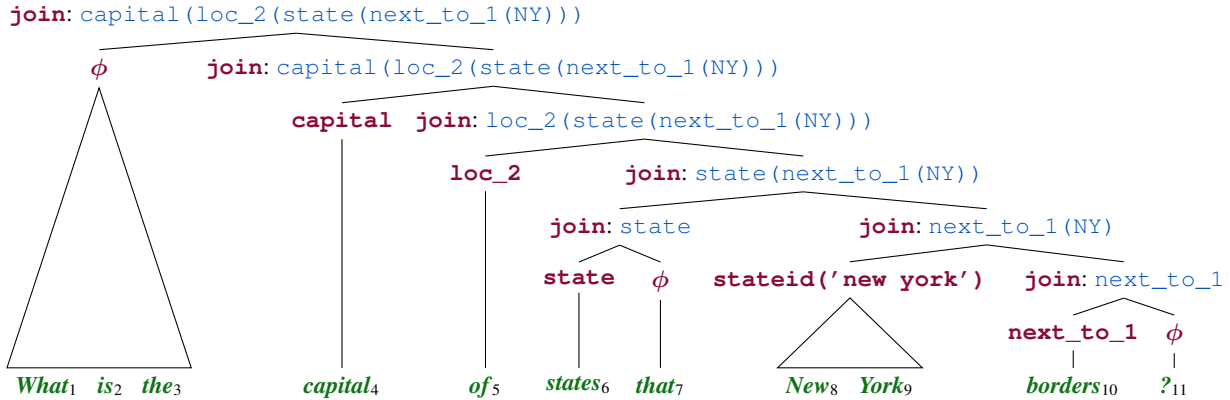

Despite the success of seq2seq models, recently, \newcite{finegan-dollak-etal-2018-improving} and \newcite{keysers2020measuring}
and \newcite{herzig-berant-2019-dont}
demonstrated that such models fail at
\emph{compositional generalization}, that is, they do not generalize to program structures that were not seen at training time. For example, a model that observes at training time the questions \utt{What states border China?} and \utt{What is the largest state?} fails to generalize to questions such as \utt{What states border the largest state?}. This is manifested in large performance drops on data splits designed to measure compositional generalization (\emph{compositional splits}), and is in contrast to the generalization abilities of humans \cite{fodor1988}.

In this work, we posit that the poor generalization of seq2seq models is  due to fact that the input utterance and output program are only tied \emph{softly} through  attention. We revisit a more traditional approach for semantic parsing \cite{zelle1996learning,10.5555/3020336.3020416,liang-etal-2011-learning},
where partial programs are predicted over short spans in the utterance, and are composed to build the program for the entire utterance. Such explicit inductive bias for compositionality should encourage compositional generalization.

Specifically, we propose to introduce such inductive bias via a span-based parser \cite{stern-etal-2017-minimal,pasupat-etal-2019-span}, equipped with the advantages of modern neural architectures. Our model, \sbm, predicts for every span in the input \emph{a category}, which is either a constant from the underlying knowledge-base, a composition category, or a null category. Given the category predictions for all spans, we can construct a tree over the input utterance and deterministically compute the output program.
For example, in Figure \ref{fig:tree_example}, the category for the tree node covering the span \utt{New York borders ?} is the composition category \s, indicating the composition of the predicate \texttt{next\_to\_1} with the entity \texttt{stateid('new york')}.

Categories are predicted for each span independently, resulting in a very simple training procedure. CKY is used at inference time to find the best \emph{span tree}, which is a tree with a category predicted at every node. The output program is computed from this tree in a bottom-up manner.

We enhance the applicability of span-based semantic parsers \cite{pasupat-etal-2019-span} in terms of both \emph{supervision} and \emph{expressivity}, by overcoming two technical challenges. First, we do not use gold trees as supervision, only programs with no explicit decomposition over the input utterance. To train with latent trees, we use a hard-EM approach, where we search for the best tree under the current model corresponding to the gold program, and update the model based on this tree. Second, some gold trees are non-projective, and cannot be parsed with a binary grammar. Thus, we extend the grammar of CKY to capture a class of non-projective structures that are common in semantic parsing. This leads to a model that is comparable and competitive with the prevailing seq2seq approach.

We evaluate our approach on three datasets, and find that \sbm
performs similarly to strong seq2seq baselines on standard i.i.d (random) splits, but dramatically improves performance on compositional splits, by 32.9, 34.6 and 13.5 absolute accuracy points on \geo \cite{zelle1996learning}, \closure \cite{bahdanau2019closure}, and \scan \cite{lake-2018-scan} respectively. Our code and data are available at \href{https://github.com/jonathanherzig/span-based-sp}{https://github.com/jonathanherzig/span-based-sp}.

\section{Problem Setup}
\label{sec:overview}

We define span-based semantic parsing as follows. Given a training set $\{(x^i,z^i)\}_{i=1}^{M}$, where $x_i$ is an utterance and $z_i$ is the corresponding program, our goal is to learn a model that maps a new utterance $x$ to a \emph{span tree} $T$ (defined below), such that \texttt{program($T$)}$=z$. The deterministic function \texttt{program($\cdot$)} maps span trees to programs.

\paragraph{Span trees} A span tree  $T$ is a tree (see Figure~\ref{fig:tree_example}) where, similar to constituency trees, each node covers a span $(i, j)$ with tokens $x_{i:j}=(x_i, x_{i+1}, \dots , x_j)$. A span tree can be viewed as a mapping from every span $(i, j)$ to a single category $c \in \mathcal{C}$, where categories describe how the meaning of a node is derived from the meaning of its children. A category $c$ is one of the following:
\vspace{-2mm}
\begin{itemize}[leftmargin=*,itemsep=0pt]
    \item \kbc : a set of domain-specific categories
    representing domain constants, including entities and predicates. E.g., in Figure \ref{fig:tree_example}, \texttt{capital}, \texttt{state}, \texttt{loc\_2} and \texttt{next\_to\_1} are binary predicates, and \texttt{stateid('new york')} is an entity. \vspace{-1.5mm}
    \item \s: a category for a node whose meaning is derived from the meaning of its two children. At most one of the children's categories can be the \ns category. \vspace{-1.5mm}
    \item \ns : a category for (i) a 
    node that does not affect the meaning of the utterance. For example, in Figure \ref{fig:tree_example}, the nodes that cover \utt{What is the} and \utt{?} are tagged by \ns; (ii) spans that do not correspond to constituents (tree nodes).
\end{itemize}
\vspace{-1.5mm}

Overall, the category set is $\mathcal{C} = \Sigma \cup \{\ns, \s \}$. We also define the \emph{terminal nodes set} $\Sigma^+=\Sigma \cup \{\ns\}$, corresponding to categories that are directly over the utterance.

\paragraph{Computing programs for span trees}
Given a mapping from spans to categories specifying a span tree $T$, we use the function $\texttt{program}(\cdot)$ to find the program for $T$. Concretely, $\texttt{program}(T)$ iterates over the nodes in $T$ bottom-up, and generates a program $z_{i:j}$ for each node covering the span $(i,j)$.

The program $z_{i:j}$ is computed deterministically. For a node with category $c \in \kbc$, $z_{i:j} = c$.  For a \s node over the span $(i,j)$, we determine $z_{i:j}$  by composing the programs of its children, $z_{i:s}$ and $z_{s,j}$ where $s$ is the \emph{split point}. As in Combinatory Categorical Grammar \cite{steedman2000syntactic}, composition is simply function application, where a domain-specific type system is used to determine which child is the function and which is the argument (along with the exact argument position for predicates with multiple arguments). If the category of one of the children is \ns, the program for $z_{i:j}$ is copied from the other child.
E.g., in Figure \ref{fig:tree_example}, the span $(8,9)$, where $z_{8:9}=\texttt{stateid('new york')}$ combines with the span $(10,11)$, where $z_{10:11}=\texttt{next\_to\_1}$. As $z_{10:11}$ is a binary predicate that takes an argument of type \texttt{state}, and $z_{8:9}$ is an entity of type \texttt{state}, the output program is $z_{8:11}=\texttt{next\_to\_1(stateid('new york'))}$.
If no combination is possible according to the type system, the execution of $\texttt{program}(T)$ fails (\S\ref{ssec:inference}).

Unlike seq2seq models, computing programs with span trees is explicitly compositional. Our main hypothesis is that this strong inductive bias should improve compositional generalization.
\section{A Span-based Semantic Parser}

Span-based parsing had success in both syntactic \cite{stern-etal-2017-minimal,kitaev-klein-2018-constituency} and semantic parsing \cite{pasupat-etal-2019-span}. The intuition is that modern sequence encoders are powerful, and thus we can predict a category for every span \emph{independently}, reducing the role of global structure. This leads to simple and fast training. 

Specifically, our parser is based on a model $p_\theta(T[i, j] = c)$, parameterized by $\theta$, that provides for every span $(i,j)$ a distribution over categories $c \in \mathcal{C}$. Due to the above independence assumption, the log-likelihood of a tree $T$ is defined as:
\begin{align}
\label{eq:tree_ll}
\log p(T)=\sum_{i<j}\log p_\theta(T[i, j]),
\end{align}
where, similar to \newcite{pasupat-etal-2019-span}, the sum is over \emph{all} spans $i<j$ and not only over constituents. We next describe the model $p_\theta(T[i, j])$ and its training, assuming we have access to gold span trees at training time (\S\ref{ssec:model}). We will later (\S\ref{ssec:weak_train}) remove this assumption, and describe a CKY-based inference procedure (\S\ref{ssec:inference}) that finds for every training example $(x,z)$ the (approximately) most probable span tree $T_{\text{train}}^*$, such that $\texttt{program}(T_{\text{train}}^*)=z$. 
We use $T_{\text{train}}^*$ as a replacement for the gold tree.
Last, we present an extension of our model that covers a class of span trees that are non-projective (\S\ref{ssec:non-projective}).

\subsection{Model}

\label{ssec:model}

We describe the architecture and training procedure of our model (\sbm), assuming we are given for every utterance $x$ a gold tree $T$, for which  $\texttt{program(T)}=z$.

Similar to \newcite{pasupat-etal-2019-span}, we minimize the negative log-likelihood $-\log p(T)$ (Eq. \ref{eq:tree_ll}) for the gold tree $T$. The loss decomposes over spans into cross-entropy terms for every span $(i, j)$. This effectively results in multi-class problem, where for every span $x_{i:j}$ we predict a category $c \in \mathcal{C}$. Training in this setup is trivial and does not require any structured inference.

Concretely, the architecture of \sbm is based on a \bb encoder \cite{devlin-etal-2019-bert} that yields a contextual representation $\mathbf{h}_i \in \mathbb{R}^{h_{\text{dim}}}$ for each token $x_i$ in the input utterance. We represent each span $(i, j)$ by concatenating its start and end representations $[\mathbf{h}_i ; \mathbf{h}_j]$, and apply a 1-hidden layer network to produce a real-valued score $\texttt{s}(x_{i:j}, c)$ for a span $(i,j)$ and category $c$:
\begin{align}
   \texttt{s}(x_{i:j}, c) = [\mathbf{W}_2\text{relu}(\mathbf{W}_1[\mathbf{h}_i ; \mathbf{h}_j])]_{\texttt{ind}(c)},
\label{eq:logits}
\end{align}
where $\mathbf{W}_1 \in \mathbb{R}^{250 \times 2h_{\text{dim}}}$, $\mathbf{W}_2 \in \mathbb{R}^{|\mathcal{C}| \times 250}$, and $\texttt{ind}(c)$ is the index of the category $c$. We take a softmax to produce the probabilities:
\begin{align}
    p_\theta(T[i, j]=c)=\frac{\text{exp}[\texttt{s}(x_{i:j}, c)]}{\sum_{c'} \text{exp}[\texttt{s}(x_{i:j}, c')]},
\label{eq:prob}
\end{align}
and train the model with a cross-entropy loss averaged over all spans, as mentioned above.

\subsection{CKY-based Inference}
\label{ssec:inference}

While we assume span-independence at training time, at test time we must output a valid span tree. We now describe an approximate $K$-best CKY algorithm that searches for the $K$ most probable trees under $p(T)$, and returns the highest-scoring one that is \emph{semantically valid}, i.e., that can be mapped to a program.\footnote{The requirement that trees are semantically valid is what prevents exact search.}
As we elaborate below, some trees cannot be mapped to a program, due to violations of the type system.

We start by re-writing our objective function, as proposed in \newcite{pasupat-etal-2019-span}. Given our definition for $p_\theta(T[i, j]=c)$, the log-likelihood is:
\begin{align*}
\label{eq:ll_deriv}
\begin{split}
\log p(T)=\sum_{i<j}\log p_{\theta}(T[i, j])= \\ \sum_{i<j} \left[ \texttt{s}(x_{i:j}, T[i, j]) - \log \sum_{c'} \text{exp}[\texttt{s}(x_{i:j}, c')] \right].
\end{split}
\end{align*}
We shift the scoring function $\texttt{s}(\cdot)$ for each span, such that the score for the \ns category is zero:
\begin{align*}
\texttt{s'}(x_{i:j}, \cdot) \vcentcolon= \texttt{s}(x_{i:j}, \cdot) - \texttt{s}(x_{i:j}, \ns).
\end{align*}
Because softmax is shift-invariant, we can replace
$\texttt{s}(\cdot)$ for $\texttt{s}'(\cdot)$ 
and preserve correctness.
This is motivated by the fact that \ns nodes, such as the one covering \utt{What is the} in Figure~\ref{fig:tree_example}, do not affect the semantics of utterance.
By shifting scores such that for all spans $\texttt{s'}(x_{i:j}, \ns)=0$, their score does not affect the overall tree score. Spans that do not correspond to tree nodes are labeled by \ns and also do not affect the tree score.

\begin{figure}[t]
\centering

  {
	\setlength{\fboxsep}{4pt}
	\fbox{
		\parbox{0.7\columnwidth}{

% 	\small
	%\textbf{CKY grammar rules}
	
    \sent $\vcentcolon=$ \s \s | \ns \s
	
    \s $\vcentcolon=$ \s \s | \s \ns

		}}
  }
  \caption{CKY grammar defining the possible output trees.}
	\label{fig:grammar}
\end{figure}

Furthermore, as $\sum_{i < j}\log \sum_{c'} \text{exp}[\texttt{s'}(x_{i:j}, c')]$ does not depend on $T$ at all, maximizing $\log p(T)$ is equivalent to maximizing the tree score:

\begin{align*}
\texttt{S}(T)\vcentcolon=\sum_{i<j}\texttt{s'}(x_{i:j}, T[i, j]).
\end{align*}
This scoring function can be maximized using CKY \cite{cocke1969,kasami1965,younger1967}. We now propose a grammar, which imposes further restrictions on the space of possible output trees at inference time.

We use a small grammar $G = (N,\Sigma^+,R,S)$, where $N=\{\texttt{S}, \s \}$ 
is the set of non-terminals, $\Sigma^+$ is the set of terminals (defined in \S\ref{sec:overview}), $R$ is a set of four rules detailed in Figure \ref{fig:grammar}, and $S$ is a special 
start symbol. 
The four grammar rules impose the following constraints on the set of possible output trees: (a) a \s or $S$ node can have at most one \ns child, as explained in \S\ref{sec:overview}; (b) nodes with no semantics combine with semantic elements on their left; (c) except at the root where they combine with elements on their right. Imposing such consistent tree structure is useful for training \sbm when predicted trees are used for training (\S\ref{ssec:weak_train}). 

The grammar $G$ can generate trees that are not semantically valid.
For example, we could generate the program \texttt{capital(placeid('mount mckinley'))}, which is semantically vacuous. We use a domain-specific type system and assign the score $\texttt{S}(T)=-\infty$ to every tree that yields a semantically invalid program. This global factor prevents exact inference, and thus we perform $K$-best parsing, keeping the top-$K$ ($K=5$) best trees for every span $(i,j)$ and non-terminal.

\begin{algorithm}[t]
\SetInd{0.25em}{0.1em}
\footnotesize
\SetKwInput{KwInput}{Input}                % Set the Input
\SetKwInput{KwOutput}{Output}              % set the Output
\DontPrintSemicolon
  
  \KwInput{$\forall i,j,c:  s(x_{i:j}, c), G =(N,\Sigma^+,R,S), x$}
  \KwOutput{$\pi$ - scores for each span and non-terminal}

        \For{$1 \leq i \leq  j \leq |x|$ } { \label{line:init_start}
            $\pi(i,j, \s) \gets \max\limits_{c \in \Sigma}\texttt{s'}(x_{ij}, c)$  \;
            $\pi(i,j, \ns) \gets \texttt{s'}(x_{ij}, \ns)$\ \tcp*[f]{equals zero}\; \label{line:init_end}
        }
        \For{$1 \leq i \leq  j \leq |x|$}{
            \For{$X \in N$}{
                $\text{temp} \gets \max\limits_{\substack{(X \rightarrow YZ) \in R \\ s \in i \dots (j-1)}} [\texttt{s'}(x_{ij}, \s) + \pi(i,s, Y) + \pi(s+1,j, Z)] $ \label{line:composition}\;
                $\pi(i,j,X) \gets \max\limits(\text{temp}, \pi(i,j, \s))$ \label{line:merge}
            }
        }
        \KwRet $\pi$\;
\caption{CKY inference algorithm}\label{alg:cky}
\end{algorithm}

Alg.~\ref{alg:cky} summarizes CKY inference, that outputs $\pi(i,j,X)$, the maximal score for a tree with non-terminal root $X$ over the span $(i,j)$. 
In Lines~\ref{line:init_start}-\ref{line:init_end} we initialize the parse chart, by going over all spans and setting $\pi(i,j,\s)$  
to the top-$K$ highest scoring domain constants ($\Sigma$), and fixing the score for \ns to be zero. We then perform the typical CKY recursion to find the top-$K$ trees that can be constructed through composition (Line~\ref{line:composition}), merge them with the domain constants found during initialization (Line~\ref{line:merge}), and keep the overall top-$K$ trees. 

Once inference is done, we retrieve the top-$K$ trees from $\pi(1, |x|, S)$, iterate over them in descending score order, and return the first tree $T^*$ that is semantically valid.

\subsection{Training without Gold Trees}
\label{ssec:weak_train}

We now remove the assumption of access to gold trees at training time, in line with standard supervised semantic parsing, where only the gold program $z$ is given, without its decomposition over $x$. 
This can be viewed as a \emph{weakly-supervised} setting, where the correct span tree is a discrete latent variable.
In this setup, our goal is to maximize 
\begin{align*}
\log p(z \mid x)&=\log \sum_{T : program(T) = z} p(T) \\
&\approx \log \argmax_{T : program(T) = z} p(T).
%\label{eq:hard_em}
\end{align*}
Because marginalizing over trees is intractable, we take a hard-EM approach \cite{liang2017nsm, min-etal-2019-discrete}, and replace the sum over trees with an $\argmax$.
More concretely, to approximately solve the $\argmax$ and find the highest scoring tree, $T_{\text{train}}^*$, we employ a constrained version of Alg. \ref{alg:cky}, that prunes out trees that cannot generate $z$.%the gold program $z$.

We first remove all predictions of constants that do not appear in $z$ by setting their score to $- \infty$: 
\begin{align*}
 \forall c \in \{\kbc \setminus \texttt{const}(z)\},i,j: s'(x_{i:j}, c) \vcentcolon= - \infty,  
\end{align*}
where $\texttt{const}(z)$ is the set of domain constants appearing in $z$. Second, we allow a composition of two nodes covering spans $(i,s)$ and $(s,j)$ only if their sub-programs $z_{i:s}$ and $z_{s:j}$ can compose according to $z$. For instance, in Figure \ref{fig:tree_example}, a span with the sub-program \texttt{capital} can only compose with a span with the sub-program  \texttt{loc\_2($\cdot$)}.
After running this constrained CKY procedure we return the highest scoring tree that yields the correct program, $T_{\text{train}}^*$, if one is found.
We then treat the span structure of $T_{\text{train}}^*$ as labels for training the parameters of \sbm.

Past work on weakly-supervised semantic parsing often used maximum marginal likelihood, especially when training from denotations only 
\cite{guu-etal-2017-language}.
In this work, we found hard-EM to be simple and sufficient, since we are given the program $z$ that provides a rich signal for guiding search in the space of latent trees.

\paragraph{Exact match features}
The challenge of weakly-supervised parsing is that \sbm must learn to map language phrases to constants, and how the span tree is structured. To alleviate the language-to-constant problem we add an exact match feature, based on a small lexicon, indicating whether a phrase in $x$ matches the language description of a category $c \in \kbc$.
These features are considered in \sbm when some phrase matches a category from \kbc, updating the score $\texttt{s}(x_{i:j}, c)$ to be: $[\mathbf{W}_2\text{relu}(\mathbf{W}_1[\mathbf{h}_i ; \mathbf{h}_j])]_{\texttt{ind}(c)} + \lambda\delta(x_{i:j},c)$,
where $\delta(x_{i:j},c)$ is an indicator that returns $1$ if $c \in \texttt{lexicon}[x_{i:j}]$, and $0$ otherwise, and $\lambda$ is a hyper-parameter that sets the feature's importance.

We use two types of $\texttt{lexicon}[\cdot]$ functions. In the first, the lexicon is created automatically to map the names of entities (not predicates), as they appear in \kbc, to their corresponding constant (e.g., $\texttt{lexicon}[\utt{new york}]=\texttt{stateid('new york')}$). This endows \sbm with a \emph{copying mechanism},  similar to seq2seq models, for predicting entities unseen during training.  
In the second lexicon we manually add no more than two examples of language phrases for each constant in \kbc. E.g., for the predicate \texttt{next\_to\_1}, we update the lexicon to include $\texttt{lexicon}[\utt{border}]=\texttt{lexicon}[\utt{borders}]=\texttt{next\_to\_1}$.
This requires minimal manual work (if no language phrases are available), 
but is done only once, and is common in semantic parsing \cite{10.5555/3020336.3020416,wang-etal-2015-building,liang2017nsm}.

\subsection{Non-Projective Trees}
\label{ssec:non-projective}

\begin{figure}[t] 
\centering
\footnotesize
\begin{tikzpicture}
\tikzset{sibling distance=-2.0pt}
\tikzset{level distance=22pt}
\tikzset{frontier/.style={distance from root=89pt}}

\Tree [.\node(AA) {\tnode{\s{}}:\;\tprog{largest\_one(pop\_1(state(all)))}}; [.- [.\tnode{\s{}}:\;\tprog{state} [.\tnode{state} \ttok{State}{1} ] [.\tnode{\nsb{}} \edge[roof]; {\ttok{that}{2}\; \ttok{has}{3}\; \ttok{the}{4}}\; ] ] \edge[white]; [.\node(BB) {\tnode{largest\_one}}; \ttok{most}{5} ] [.\tnode{\s{}}:\;\tprog{pop\_1} [.\tnode{pop\_1} \ttok{people}{6} ] [.\tnode{\nsb{}} \ttok{?}{7} ] ] ] ]
\draw [dashed] (AA.south) --(BB.north);
\end{tikzpicture}
\caption{An example of a non-projective tree. The corresponding program $z$ is at the root.}
\label{fig:non_proj}
\end{figure}
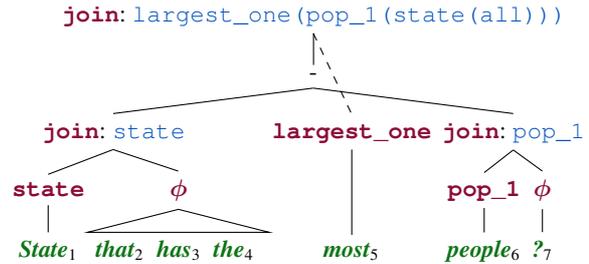

Our span-based parser assumes composition can only be done for adjacent spans that form together a contiguous span. However, this assumption does not always hold \cite{liang-etal-2011-learning}. 
For example, %for the utterance 
in Figure \ref{fig:non_proj}, while the predicate \texttt{pop\_1} should combine with the predicate \texttt{state}, the spans they align to 
%in the utterance 
(\utt{people} and \utt{state} respectively) are not contiguous, as they are separated by \utt{most}, which contributes the semantics of a superlative.

In constituency parsing, such non-projective structures are treated by adding rules to the grammar $G$ \cite{maier-etal-2012-plcfrs,corro2020span,stanojevic-steedman-2020-span}. We identify one specific class of non-projective structures that is frequent in semantic parsing (Figure \ref{fig:non_proj}), and expand the grammar $G$ and the CKY Algorithm to support this structure.
Specifically, we add the ternary grammar rule $\s \vcentcolon= \s\;\s\;\s$. During CKY, when calculating the top-$K$ trees for spans $(i,j)$ (line~\ref{line:composition} in Alg. \ref{alg:cky}), we also consider the following top-$K$ scores for the non-terminal \s:
\begin{align*}
     &\max\limits_{\substack{s_1 \in i \dots (j-2) \\ s_2 \in (s_1+1) \dots (j-1)}} [\texttt{s'}(x_{ij},\s) + \pi(i,s_1, \s) \\ &+ \pi(s_1+1,s_2,\s) + \pi(s_2+1,j,\s)].
\end{align*}
These additional trees are created by going over all possible ways of dividing a span $(i,j)$ into three parts. The score of the sub-tree is then the sum of the score of the root added to the scores of the three children. To compute the program for such ternary nodes, we again use our type system, where we first compose the programs of the two outer spans $(i,s_1)$ and $(s_2+1, j)$ and then compose the resulting program with the program corresponding to the span $(s_1+1,s_2)$. Supporting ternary nodes in the tree increases the time complexity of CKY from $O(n^3)$ to $O(n^4)$ for our implementation.\footnote{\newcite{corro2020span} show an $O(n^3)$ algorithm for this type of non-projective structure.}

\section{Experiments and Results}

\begin{table}[t]
\centering
\resizebox{1.0\columnwidth}{!}{
\begin{tabular}{llccc}
\toprule
\textbf{Dataset} & \textbf{Split} & \textbf{train} & \textbf{dev} & \textbf{test} \\
\midrule
\multirow{3}{*}{\textsc{\scansp}} & \iid  & 13,383     & 3,345 & 4,182 \\
                                  & \textsc{right}  & 12,180      & 3,045 & 4,476  \\
                                  & \textsc{aroundRight}  & 12,180      & 3,045 & 4,476  \\ \midrule
\multirow{2}{*}{\textsc{\clevr}} & \iid  & 694,689     & 5,000 & 149,991 \\
                                  & \closure  & 694,689     & 5,000 & 25,200  \\ \midrule
\multirow{3}{*}{\textsc{\geo}} & \iid  & 540     & 60 & 280 \\
                                  & \textsc{template}  & 544      & 60 & 276  \\
                                  & \textsc{length}  & 540      & 60 & 280  \\
\bottomrule
\end{tabular}}
\caption{\label{tab:stats} Number of examples for all datasets.}
\end{table}

\begin{table*}[tb]
\begin{center}
\resizebox{1.0\textwidth}{!}{
\begin{tabular}{@{}lllllllllllllllll@{}}
\toprule
\multirow{3}{*}{\textbf{Model}} & \multicolumn{6}{c}{\textbf{\scansp}}                                                                                                                                        & \multicolumn{4}{c}{\textbf{\clevr}}                                                                                 & \multicolumn{4}{c}{\textbf{\geo}}                                                                                   \\
\cmidrule(lr){2-7}
\cmidrule(lr){8-11}
\cmidrule(lr){12-17}
                      & \multicolumn{2}{c}{\textbf{\iid}}                             & \multicolumn{2}{c}{\textsc{\textbf{right}}}                           & \multicolumn{2}{c}{\textsc{\textbf{aroundRight}}}                    & \multicolumn{2}{c}{\textbf{\iid}}                             & \multicolumn{2}{c}{\textbf{\closure}}                         & \multicolumn{2}{c}{\textbf{\iid}}                             & \multicolumn{2}{c}{\textsc{\textbf{template}}}
                                                & \multicolumn{2}{c}{\textsc{\textbf{length}}}\\
\cmidrule(lr){2-3}
\cmidrule(lr){4-5}
\cmidrule(lr){6-7}
\cmidrule(lr){8-9}
\cmidrule(lr){10-11}
\cmidrule(lr){12-13}
\cmidrule(lr){14-15}
\cmidrule(lr){16-17}
                      & \multicolumn{1}{c}{\textbf{dev}} & \multicolumn{1}{c}{\textbf{test}} & \multicolumn{1}{c}{\textbf{dev}} & \multicolumn{1}{c}{\textbf{test}} & \multicolumn{1}{c}{\textbf{dev}} & \multicolumn{1}{c}{\textbf{test}} & \multicolumn{1}{c}{\textbf{dev}} & \multicolumn{1}{c}{\textbf{test}} & \multicolumn{1}{c}{\textbf{dev}} & \multicolumn{1}{c}{\textbf{test}} & \multicolumn{1}{c}{\textbf{dev}} & \multicolumn{1}{c}{\textbf{test}} & \multicolumn{1}{c}{\textbf{dev}} & \multicolumn{1}{c}{\textbf{test}}
                      & \multicolumn{1}{c}{\textbf{dev}} & \multicolumn{1}{c}{\textbf{test}}
                      \\
\cmidrule(lr){2-2}
\cmidrule(lr){3-3}
\cmidrule(lr){4-4}
\cmidrule(lr){5-5}
\cmidrule(lr){6-6}
\cmidrule(lr){7-7}
\cmidrule(lr){8-8}
\cmidrule(lr){9-9}
\cmidrule(lr){10-10}
\cmidrule(lr){11-11}
\cmidrule(lr){12-12}
\cmidrule(lr){13-13}
\cmidrule(lr){14-14}
\cmidrule(lr){15-15}
\cmidrule(lr){16-16}
\cmidrule(lr){17-17}
\stos                      & \multicolumn{1}{c}{100} & \multicolumn{1}{c}{99.9} & \multicolumn{1}{c}{100} & \multicolumn{1}{c}{11.6} & \multicolumn{1}{c}{100} & \multicolumn{1}{c}{0.0} & \multicolumn{1}{c}{100} & \multicolumn{1}{c}{100} & \multicolumn{1}{c}{100} & \multicolumn{1}{c}{59.5} & \multicolumn{1}{c}{83.3} & \multicolumn{1}{c}{78.5} & \multicolumn{1}{c}{71.6} & \multicolumn{1}{c}{46.0} & \multicolumn{1}{c}{86.7} & \multicolumn{1}{c}{24.3}\\
\addlinespace[3pt]
\;\;+ELMo                      & \multicolumn{1}{c}{100} & \multicolumn{1}{c}{100} & \multicolumn{1}{c}{100} & \multicolumn{1}{c}{54.9} & \multicolumn{1}{c}{100} & \multicolumn{1}{c}{41.6} & \multicolumn{1}{c}{100} & \multicolumn{1}{c}{100} & \multicolumn{1}{c}{100} & \multicolumn{1}{c}{64.2} & \multicolumn{1}{c}{83.3} & \multicolumn{1}{c}{79.3} & \multicolumn{1}{c}{83.3} & \multicolumn{1}{c}{50.0} & \multicolumn{1}{c}{86.7} & \multicolumn{1}{c}{25.7}\\
\addlinespace[3pt]
\btos                      & \multicolumn{1}{c}{99.9} & \multicolumn{1}{c}{100} & \multicolumn{1}{c}{99.9} & \multicolumn{1}{c}{77.7} & \multicolumn{1}{c}{99.9} & \multicolumn{1}{c}{95.3} & \multicolumn{1}{c}{100} & \multicolumn{1}{c}{100} & \multicolumn{1}{c}{100} & \multicolumn{1}{c}{56.4} & \multicolumn{1}{c}{88.3} & \multicolumn{1}{c}{81.1} & \multicolumn{1}{c}{85.0} & \multicolumn{1}{c}{49.6} & \multicolumn{1}{c}{90.0} & \multicolumn{1}{c}{26.1}\\
\addlinespace[3pt]
\grammar                      & \multicolumn{1}{c}{100} & \multicolumn{1}{c}{100} & \multicolumn{1}{c}{100} & \multicolumn{1}{c}{0.0} & \multicolumn{1}{c}{100} & \multicolumn{1}{c}{4.2} & \multicolumn{1}{c}{100} & \multicolumn{1}{c}{100} & \multicolumn{1}{c}{100} & \multicolumn{1}{c}{51.3} & \multicolumn{1}{c}{78.3} & \multicolumn{1}{c}{72.1} & \multicolumn{1}{c}{76.7} & \multicolumn{1}{c}{54.0} & \multicolumn{1}{c}{81.7} & \multicolumn{1}{c}{24.6}\\
\addlinespace[3pt]
\bart                      & \multicolumn{1}{c}{100} & \multicolumn{1}{c}{100} & \multicolumn{1}{c}{100} & \multicolumn{1}{c}{50.5} & \multicolumn{1}{c}{100} & \multicolumn{1}{c}{100} & \multicolumn{1}{c}{100} & \multicolumn{1}{c}{100} & \multicolumn{1}{c}{100} & \multicolumn{1}{c}{51.5} & \multicolumn{1}{c}{93.3} & \multicolumn{1}{c}{87.1} & \multicolumn{1}{c}{86.7} & \multicolumn{1}{c}{67.0} & \multicolumn{1}{c}{90.0} & \multicolumn{1}{c}{19.3}\\
\addlinespace[3pt]
\etoe                     & \multicolumn{1}{c}{-} & \multicolumn{1}{c}{-} & \multicolumn{1}{c}{-} & \multicolumn{1}{c}{-} & \multicolumn{1}{c}{-} & \multicolumn{1}{c}{-} & \multicolumn{1}{c}{99.9} & \multicolumn{1}{c}{99.8} & \multicolumn{1}{c}{99.9} & \multicolumn{1}{c}{63.3} & \multicolumn{1}{c}{-} & \multicolumn{1}{c}{-} & \multicolumn{1}{c}{-} & \multicolumn{1}{c}{-} & \multicolumn{1}{c}{-} & \multicolumn{1}{c}{-}\\
\addlinespace[3pt]
\midrule
\sbm                      & \multicolumn{1}{c}{100} & \multicolumn{1}{c}{100} & \multicolumn{1}{c}{100} & \multicolumn{1}{c}{\textbf{100}} & \multicolumn{1}{c}{100} & \multicolumn{1}{c}{\textbf{100}} & \multicolumn{1}{c}{97.0} & \multicolumn{1}{c}{96.7} & \multicolumn{1}{c}{98.9} & \multicolumn{1}{c}{\textbf{98.8}} & \multicolumn{1}{c}{88.3} & \multicolumn{1}{c}{86.1} & \multicolumn{1}{c}{93.3} & \multicolumn{1}{c}{\textbf{82.2}} & \multicolumn{1}{c}{95.0} & \multicolumn{1}{c}{\textbf{63.6}}\\
\;\;-lexicon                    & \multicolumn{1}{c}{100} & \multicolumn{1}{c}{100} & \multicolumn{1}{c}{100} & \multicolumn{1}{c}{\textbf{100}} & \multicolumn{1}{c}{100} & \multicolumn{1}{c}{\textbf{100}} & \multicolumn{1}{c}{99.4} & \multicolumn{1}{c}{99.3} & \multicolumn{1}{c}{98.5} & \multicolumn{1}{c}{88.6} & \multicolumn{1}{c}{88.3} & \multicolumn{1}{c}{78.9} & \multicolumn{1}{c}{86.7} & \multicolumn{1}{c}{65.9} & \multicolumn{1}{c}{90.0} & \multicolumn{1}{c}{41.4}\\
\;\;-non projective                       & \multicolumn{1}{c}{-} & \multicolumn{1}{c}{-} & \multicolumn{1}{c}{-} & \multicolumn{1}{c}{-} & \multicolumn{1}{c}{-} & \multicolumn{1}{c}{-} & \multicolumn{1}{c}{-} & \multicolumn{1}{c}{-} & \multicolumn{1}{c}{-} & \multicolumn{1}{c}{-} & \multicolumn{1}{c}{85.0} & \multicolumn{1}{c}{80.0} & \multicolumn{1}{c}{90.0} & \multicolumn{1}{c}{80.2} & \multicolumn{1}{c}{93.3} & \multicolumn{1}{c}{59.3}\\
\midrule
\midrule
\;\;+gold trees                      & \multicolumn{1}{c}{100} & \multicolumn{1}{c}{100} & \multicolumn{1}{c}{100} & \multicolumn{1}{c}{100} & \multicolumn{1}{c}{100} & \multicolumn{1}{c}{100} & \multicolumn{1}{c}{100} & \multicolumn{1}{c}{96.8} & \multicolumn{1}{c}{100} & \multicolumn{1}{c}{96.7} & \multicolumn{1}{c}{91.2} & \multicolumn{1}{c}{86.4} & \multicolumn{1}{c}{100} & \multicolumn{1}{c}{81.8} & \multicolumn{1}{c}{96.7} & \multicolumn{1}{c}{68.6}\\
\bottomrule
\end{tabular}}
\end{center}
\caption{\label{tab:main_res} Denotation accuracy for all models, including \sbm ablations. For both \clevr splits, \sbm only trains on $10K$ examples, in comparison to $695K$ for the baselines.}
\end{table*}

We now present our experimental evaluation, which demonstrates the advantage of span-based parsing for compositional generalization. We compare to baseline models over two types of data splits: (a) \emph{\iid split}, where the training and test sets are sampled from the same distribution, and (b) \emph{compositional split}, where the test set includes structures that are unseen at training time. Details on the experimental setup are given in Appendix~\ref{app:exp_setup}.

\subsection{Datasets}

We evaluate on the following datasets (Table \ref{tab:stats}). 

\paragraph{\geo} 
Contains $880$ questions about US  geography \cite{zelle1996learning},  using the FunQL formalism \cite{10.5555/1619499.1619504}. 
For the \iid{} split, we use the standard train/test split, randomly sampling 10\% of the training set for development. 
We additionally use two compositional splits based on program templates (\textsc{template}) and on program lengths (\textsc{length}).

For the compositional split, \textsc{template}, we use the procedure from \newcite{finegan-dollak-etal-2018-improving} and split the $880$ examples by templates. A template is created by anonymizing entities in the program to their type (both \texttt{stateid('new york')} and \texttt{stateid('utah')} are anonymized to \texttt{STATE}). We then split to train/development/test sets, such that all examples that share a template are assigned to the same set. We also verify that the sizes of theses sets are as close as possible to the \iid split. 

For the compositional split, \textsc{length}, we sort the dataset by program token length and take the longest 280 examples to be the test set. We then randomly split the shortest 600 examples between the train and development set, where we take 10\% of the 600 examples for the latter.

\paragraph{\clevr and \closure}
\clevr \cite{johnson2017clevr} contains synthetic questions, created using 80 templates, over synthetic images with multiple objects of different shapes, colors, materials and sizes
(example in Fig. \ref{fig:clevr} in the Appendix). The recent \closure dataset \cite{bahdanau2019closure}, includes seven new question templates that are created by combining referring expressions of various types from \clevr in new ways.

We use the semantic parsing version of these datasets, where each image is described by a \emph{scene} (knowledge-base) that holds the attributes and positional relations of all objects. We use programs in the DSL version from \newcite{Mao2019NeuroSymbolic}.

For our experiments, we take $5K$ examples from the original \clevr training set and treat them as our development set. We use the other $695K$ examples as training data for our baselines. Importantly, we \emph{only} use $10K$ training examples for \sbm to reduce training time. We then create an \iid split where we test on the \clevr original development set (test scenes are not publicly available). We additionally define the \closure split, that tests compositional generalization, where we test on \closure. 

\paragraph{\scansp}
\scan \cite{lake-2018-scan} contains natural language navigation commands that are mapped to action sequences ($x$ and $y$ in Fig. \ref{fig:scan} in the Appendix).
As \scan lacks programs, we automatically translate the input to programs ($z$ in Fig. \ref{fig:scan}) to crate the semantic parsing version of \scan, denoted \scansp (more details are given in Appendix~\ref{app:scan}). 
We experiment with the random \textsc{simple} split from \newcite{lake-2018-scan} as our \iid split. we further use the \emph{primitive right} (\textsc{right}) and \emph{primitive around right} (\textsc{aroundRight}) compositional splits from \newcite{loula-etal-2018-rearranging}.
For each split we randomly assign $20\%$ of the training set for development. 

\subsection{Baselines}

\paragraph{\stos} Similar to \newcite{finegan-dollak-etal-2018-improving}, our baseline parser is a standard seq2seq model~\cite{jia2016recombination} that encodes the utterance $x$ with a BiLSTM encoder over pre-trained GloVe~\cite{pennington2014glove} or \textsc{ELMo}~\cite{peters-etal-2018-deep} embeddings, and decodes the program with an attention-based LSTM decoder~\cite{bahdanau2015neural} assisted by a copying mechanism for handling entities unseen during training time~\cite{gu-etal-2016-incorporating}.

\paragraph{\btos}
Same as \stos, but we replace the BiLSTM encoder with \bb, which is identical to the encoder of \sbm{}.

\paragraph{\grammar}
Grammar-based decoding has been shown to improve performance on \iid splits \cite{krishnamurthy2017neural,yin-neubig-2017-syntactic}. 
Because decoding is constrained by the grammar, the model outputs only valid programs, which could potentially improve performance on compositional splits. 
We use the grammar from \cite{wong-mooney-2007-learning} for \geo, and write grammars for \scansp and \clevr + \closure. The model architecture is identical to \stos.

\paragraph{\bart}
We additionally experiment with \bartb \cite{lewis-etal-2020-bart}, a seq2seq model pre-trained as a denoising autoencoder.

\paragraph{\etoe}
Semantic parsers generate a program that is executed to retrieve an answer. However, other end-to-end models directly predict the answer from the context without an executor, where the context can be an image \cite{hudson2018compositional,perez2018film}, a table \cite{herzig-etal-2020-tapas}, etc.
Because \clevr and \closure have a closed set of 28 possible answers and a short context (the scene), they are a good fit for end-to-end approaches. To check whether end-to-end models generalize compositionally, we implement the following model. We use \bb to encode the concatenation of the input $x$ to a representation of all objects in the scene. Each scene object is represented by \emph{adding} learned embeddings of all of its attributes: shape, material, size, color, and relative positional rank (from left to right, and from front to back). We fine-tune the model on the training set using cross-entropy loss, where the \texttt{[CLS]} token is used to predict the answer.

\subsection{Main Results}

Table \ref{tab:main_res} shows denotation accuracies for all baselines (top part) and our \sbm model (middle part). For \sbm, We also ablate the use of the manually constructed lexicon (\S\ref{ssec:weak_train}) and the non-projective extension to CKY (\S\ref{ssec:non-projective}), which is relevant only for \geo, where non-projective structures are more frequent.

The table shows that all baselines generalize well on the \iid split, but suffer from a large accuracy drop on the compositional splits (except \btos and \bart on \textsc{aroundRight}). For instance, on the compositional \closure split, all baselines achieve accuracy in the range of $51.3-64.2$, while performing perfectly on the \iid split. 
Conversely, \sbm performs almost identically on both splits. \sbm attains near-perfect performance on all \scansp and \clevr splits, despite training on \emph{only} $10K$ examples from \clevr compared to $695K$ training examples for the baselines (70x less data). On \geo, \sbm performs similarly to other semantic parsers on the \iid split \cite{dong-lapata-2016-language}, and loses just $4$ points on the compositional \textsc{template} split. On the \textsc{length} split, \sbm{} yields an accuracy of 63.6, substantially outperforming all baselines by more than 37 accuracy points.  

Our ablations show that the lexicon is crucial for \geo, which has a small training set. In this setting, learning the mapping from language phrases to predicates is challenging. Ablating non-projective parsing also hurts performance for \geo, and leads to a reduction of 2-6 points for all of the splits.

\subsection{Decomposition Analysis}

We now analyze whether trees learned by \sbm are similar to gold trees. For this analysis we semi-automatically annotate our datasets with gold trees. We do this by manually creating a domain-specific lexicon for each dataset (extending the lexicon from \S\ref{ssec:weak_train}), mapping domain constants to possible phrases in the input utterances. We then, for each example, traverse the program tree (rather than the span tree) bottom-up and annotate \s and \ns categories for spans in the utterance, aided by manually-written domain-specific rules. In cases where the annotation is ambiguous, e.g., examples with more than two instances of a specific domain constant, we do not produce a gold tree.

We manage to annotate 100\%/94.9\%/95.9\% of the examples in \scansp / \geo / \clevr+ \closure respectively in this manner.
We verify the correctness of our annotation by training \sbm from our annotated gold trees (bottom part of Table \ref{tab:main_res}). The results shows that training from these ``gold'' trees leads to similar performance as training only from programs.

\begin{table}[t]
\centering
\resizebox{0.7\columnwidth}{!}{
\begin{tabular}{llc}
\toprule
\multirow{1}{*}{\textbf{Dataset}} & \multirow{1}{*}{\textbf{Split}}   & \textbf{F$_1$} \\ 
\midrule
\multirow{3}{*}{\textsc{\scansp}} & \iid  & 100       \\
                                   & \textsc{right}  & 100       \\ 
                                   & \textsc{aroundRight} & 100\\ \midrule
\multirow{2}{*}{\textsc{\clevr}}     & \iid  & 70.6     \\
                                   & \closure  & 70.6    \\ \midrule
\multirow{3}{*}{\textsc{\geo}}  & \iid  & 94.7      \\
                                   & \textsc{template}  & 91.6 \\
                                   & \textsc{length}  & 93.7 \\

\bottomrule
\end{tabular}}
\caption{\label{tab:f1} F$_1$ scores on the test set w.r.t to the semi-automatically annotated gold trees.}
\end{table}

We then train \sbm from gold programs, as explained in \S\ref{ssec:weak_train}, and calculate F$_1$ test scores, comparing the predicted span trees to the gold ones.
F$_1$ is computed between the two sets of labeled spans, taking into account both the spans and their categories, but excluding spans with the \ns category that do not contribute to the semantics. 

Table \ref{tab:f1} shows that for \geo the trees \sbm predicts are similar to the gold trees (with 94.7, 91.6 and 93.7 F$_1$ scores for the \iid, \textsc{template} and \textsc{length} splits respectively), and in \scansp we predict perfect trees. On \clevr, we get a lower F$_1$ score of 70.6 for both the \iid and \closure splits. However, when manually inspecting predicted trees on the \iid split, we notice that predicted trees that are not identical to gold trees, are actually correct. This happens in cases where multiple gold trees are possible. For instance, in Figure \ref{fig:clevr} (in the Appendix), the span $x_{13:15}=$\utt{matte block ?} can be either parsed as [matte [block ?]], as in the figure, or [[matte~ block]~ ?]. This phenomena is common in \clevr and \closure, as span trees tend to be deep, and thus have more ambiguity.

\subsection{Limitations}

Our approach assumes a one-to-one mapping between domain constants and their manifestation as phrases in language. This leads to strong results on compositional generalization, but hurts the flexibility that is sometimes necessary in semantic parsing. For example, in some cases predicates do not align explicitly to a phrase in the utterance or appear several times in the program but only once in the utterance \cite{berant-etal-2013-semantic,pasupat-liang-2015-compositional}. This is evident in text-to-SQL parsing, where an utterance such as \utt{What is the minimum, and maximum age of all singers from France?} is mapped to \texttt{SELECT min(age) , max(age) FROM singer WHERE country='France'}. Here, the constant \texttt{age} is mentioned only once in language (but twice in the program), and \texttt{country} is not mentioned at all. Thus, our approach is more suitable for formalisms where there is tighter alignment between the natural and formal language.

In addition, while we handle a class of non-projective trees (\S\ref{ssec:non-projective}), there are other non-projective structures that \sbm can not parse. Extending CKY to support all structures from \newcite{corro2020span} leads to a time complexity of $O(n^6)$, which might be impractical. 

\section{Related Work}

Until the neural era, semantic parsers used a lexicon and composition rules to predict partial programs for spans and compose them until a full program is predicted, and typically scored with a log-linear model given features over the utterance and the program
\cite{10.5555/3020336.3020416,liang-etal-2011-learning}. 
In this work, we use a similar compositional approach, but take advantage of powerful span representations based on modern neural architectures.

The most similar work to ours is by \newcite{pasupat-etal-2019-span}, who presented a neural span-based semantic parser. While they focused on training using projective gold trees (having more supervision and less expressivity than seq2seq models) and testing on i.i.d examples, we handle non-projective trees, given only program supervision, rather than trees. More importantly, we show that this approach leads to dramatic gains in compositional generalization compared to autoregressive parsers.

In recent years, work on compositional generalization in semantic parsing mainly focused on the poor performance of parsers in compositional splits \cite{finegan-dollak-etal-2018-improving}, creating new datasets that require compositional generalization \cite{keysers2020measuring,lake-2018-scan, bahdanau2019closure}, and proposing specialized architectures mainly for the \scan task \cite{meta2019blake,nye2020learning,Gordon2020Permutation,Liu2020CompositionalGB,gupta-lewis-2018-neural}. In this work we present a general-purpose architecture for semantic parsing that incorporates an inductive bias towards compositional generalization.
Finally, concurrently to us, \newcite{shaw2020compositional} induced a synchronous grammar over program and utterance pairs and used it to introduce a compositional bias, showing certain improvements over compositional splits.
\section{Conclusion}

Seq2seq models have become unprecedentedly popular in semantic parsing but struggle to generalize to unobserved structures.
In this work, we show that our span-based parser, \sbm, that precisely describes how meaning is composed over the input utterance leads to dramatic improvements in compositional generalization. % over strong seq2seq models.
%In future work, we plan to investigate hybrid models, which combine the flexibility of seq2seq models with the explicit inductive bias for compositionality inherent to span-based parsing.
In future work, we plan to investigate ways to introduce the explicit compositional bias, inherent to \sbm, directly into seq2seq models.

\section*{Acknowledgments}
We thank Ben Bogin, Nitish Gupta, Matt Gardner and the anonymous reviewers for their constructive feedback, useful comments and suggestions. This work was completed in partial fulfillment for the PhD degree of the first author, which was also supported by a Google PhD fellowship. This research was partially supported by The Yandex Initiative for Machine Learning, and the European Research Council (ERC) under the European Union Horizons 2020 research and innovation programme. (grant ERC DELPHI 802800).

\bibliographystyle{acl_natbib}
\bibliography{references}

\clearpage
\appendix\section*{Appendix}

\section{Experimental Setup}
\label{app:exp_setup}

We evaluate models with denotation accuracy, that is, the proportion of questions for which the denotations of the predicted and gold programs are identical. For \sbm, we selected a learning rate of $1e^{-5}$, considering the values $[1e^{-4},1e^{-5},1e^{-6}]$, and use a batch size of 5. For our baselines, we tune the learning rate, batch size, and dropout. We choose all hyper-parameters by early-stopping with respect to development set denotation accuracy.
Training \sbm takes between 2 hours for \geo up to 20 hours for \clevr on a single GeForce GTX 1080 GPU. 
Our seq2seq baselines are from AllenNLP \cite{gardner-etal-2018-allennlp}, and all \bb (110M parameters) implementations are from the Transformers library \cite{Wolf2019HuggingFacesTS}. 

We additionally implement executors that calculate the denotation of a program with respect to the corresponding scene for \clevr+\closure and retrieve an action sequence as the denotation for \scansp. 

\section{Generating \scansp}
\label{app:scan}
To create a semantic parsing version of \scan, we introduce the binary predicates \texttt{and}, \texttt{after}, \texttt{walk}, \texttt{jump}, \texttt{run}, \texttt{look} and \texttt{turn}. We additionally introduce the unary predicates \texttt{twice} and \texttt{thrice}. Finally, we introduce the constants \texttt{left}, \texttt{right}, \texttt{opposite} and \texttt{around}. We then construct a synchronous context-free grammar (SCFG) that parses utterances in \scan into programs in \scansp by utilizing the constants above and simple composition rules. Finally, we use our grammar to parse all utterances in \scan to generate the programs in \scansp.      

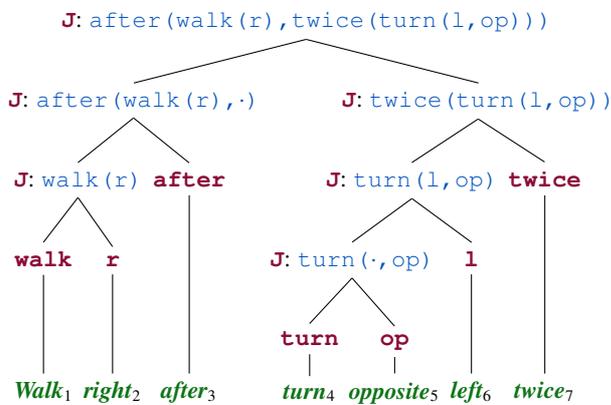
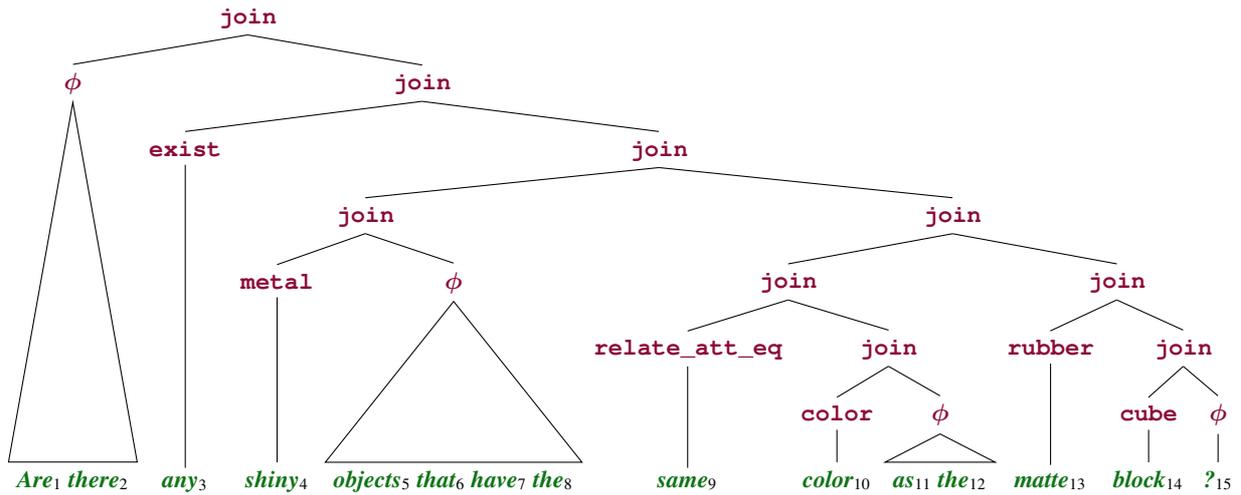
\begin{figure*}[t] 
\centering
\footnotesize
\begin{tikzpicture}
\tikzset{level distance=25pt}
\tikzset{sibling distance=1.3pt}
\tikzset{frontier/.style={distance from root=175pt}}
\node at (5,-6.8) (Sentence) {{\boldmath$x$}\textbf{:} \emph{Are there any shiny objects that have the same color as the matte block?}};
\node at (5,-7.3) (Sentence) {{\boldmath$z$}\textbf{:} \texttt{exist(filter(metal,relate\_att\_eq(color,filter(rubber,cube,scene()))))}};
\Tree [.\tnode{\s{}} 
[.\tnode{\nsb{}} \edge[roof]; {\ttok{Are}{1} \ttok{there}{2}} ] 
[.\tnode{\s{}} [.\tnode{exist} \ttok{any}{3} ] [.\tnode{\s{}} [.\tnode{\s{}} [.\tnode{metal} \ttok{shiny}{4} ] [.\tnode{\nsb{}} \edge[roof]; { \ttok{objects}{5} \ttok{that}{6} \ttok{have}{7} \ttok{the}{8}} ] ] [.\tnode{\s{}} [.\tnode{\s{}} [.\tnode{relate\_att\_eq} \ttok{same}{9} ] [.\tnode{\s{}} [.\tnode{color} \ttok{color}{10} ] [.\tnode{\nsb{}} \edge[roof]; { \ttok{as}{11} \ttok{the}{12} } ] ] ] [.\tnode{\s{}} [.\tnode{rubber} \ttok{matte}{13} ] [.\tnode{\s{}} [.\tnode{cube} \ttok{block}{14} ] [.\tnode{\nsb{}} \ttok{?}{15} ] ] ] ] ] ] ]
\end{tikzpicture}
\caption{An example span tree from \clevr, along with its utterance $x$ and program $z$. Here, the type system is used in \texttt{join} nodes to deterministically invoke the predicates \texttt{filter} and \texttt{scene} where needed. Sub-programs are omitted due to space reasons.}
\label{fig:clevr}
\end{figure*}

\begin{figure}[t] 
\centering
\footnotesize
\begin{tikzpicture}
\tikzset{sibling distance=-2.0pt}
\tikzset{frontier/.style={distance from root=140pt}}
\node at (0,-6) (Sentence) {{\boldmath$x$}\textbf{:} \emph{Walk right after turn opposite left twice}};
\node at (0,-6.5) (Sentence) {{\boldmath$z$}\textbf{:} \texttt{after(walk(r),twice(turn(l,op)))}};
\node at (0,-7) (Sentence) {{\boldmath$y$}\textbf{:} \texttt{LTURN LTURN LTURN LTURN RTURN WALK}};
\Tree [.\tnode{J}:\;\tprog{after(walk(r),twice(turn(l,op)))} [.\tnode{J}:\;\tprog{after(walk(r),$\cdot$)} [.\tnode{J}:\;\tprog{walk(r)} [.\tnode{walk} \ttok{Walk}{1} ] [.\tnode{r} \ttok{right}{2} ] ] [.\tnode{after} \ttok{after}{3} ] ] [.\tnode{J}:\;\tprog{twice(turn(l,op))} [.\tnode{J}:\;\tprog{turn(l,op)} [.\tnode{J}:\;\tprog{turn($\cdot$,op)} [.\tnode{turn} \ttok{turn}{4} ] [.\tnode{op} \ttok{opposite}{5} ] ] [.\tnode{l} \ttok{left}{6} ] ] [.\tnode{twice} \ttok{twice}{7} ] ] ]

\end{tikzpicture}
\caption{An example span tree from \scansp, along with its utterance $x$, program $z$ and action sequence $y$. The category \s is abbreviated to \texttt{J}.}
\label{fig:scan}
\end{figure}

\end{document}